\documentclass[letterpaper]{article}

\usepackage{aaai25}     
\usepackage{times}                  
\usepackage{helvet}                 
\usepackage[utf8]{inputenc} 
\usepackage{courier}                
\usepackage[hyphens]{url}           
\usepackage{graphicx}               
\usepackage{natbib}                 
\usepackage{caption}                
\usepackage{algorithm}
\usepackage[noend]{algorithmic}

\usepackage{etoolbox}       
\usepackage{amsmath}        
\usepackage{amsfonts}       
\usepackage{amssymb}        
\usepackage{soul}           
\usepackage{todonotes}      
\usepackage{dsfont}         
\usepackage{dsfont}         
\usepackage{stmaryrd}       
\usepackage{amsthm}         
\usepackage{xspace}         
\usepackage{tikz}           

\newcommand{\datad}{\mathcal{D}}
\newcommand{\inputspace}{\mathcal{X}}
\newcommand{\outputspace}{\mathcal{Y}}
\newcommand{\tpr}{TPR@$5\%$\xspace}

\DeclareMathOperator{\mudiamop}{diam}
\NewDocumentCommand{\mudiam}{O{\mu}}{\mudiamop_{#1}}

\newcommand{\baseline}{AKH\xspace}
\newcommand{\test}{\mathcal{T}}

\newcommand{\QRD}{QuRD\xspace}

\newcommand{\card}[1]{\left| #1 \right|}
\newcommand{\set}[1]{\left\{ #1 \right\}}
\newcommand{\sut}{\; : \;}

\DeclareMathOperator{\ind}{\mathds{1}}
\newcommand{\indb}[1]{\ind \left\{ #1 \right\}}
\newcommand{\snot}[1]{\overline{#1}}

\DeclareMathOperator*{\argmax}{arg\max}

\newcommand{\norm}[1]{\left\lVert #1 \right\rVert}
\newcommand{\abs}[1]{\left| #1 \right|}

\DeclareMathOperator{\proba}{\mathbb{P}}

\newcommand{\probap}[2][]{\proba_{#1} \left( #2\right)}

\newcommand{\R}{\mathbb{R}}

\newcommand{\given}{\,\middle|\,}

\newcommand{\createNameCite}[2]{%
    \newtoggle{#1}%
    \hyphenation{#1}
    \expandafter\newcommand\csname #1\endcsname{%
        \nottoggle{#1}{
            {#1}~\cite{#2}%
        }{
            #1%
        }%
        \global\toggletrue{#1}
    }%
}

\createNameCite{ModelReuse}{liModelDiffTestingbasedDNN2021}
\createNameCite{SACBench}{guanAreYouStealing2022}

\createNameCite{ZestOfLIME}{jiaZestLIMEArchitectureIndependent2022}
\createNameCite{ModelDiff}{liModelDiffTestingbasedDNN2021}
\createNameCite{SAC}{guanAreYouStealing2022}
\createNameCite{FUAP}{pengFingerprintingDeepNeural2022}
\createNameCite{FBI}{mahoFingerprintingClassifiersBenign2023}
\createNameCite{IPGuard}{caoIPGuardProtectingIntellectual2021}
\createNameCite{AFA}{zhaoAFAAdversarialFingerprinting2020}
\createNameCite{ModelGif}{songModelGiFGradientFields2023}
\createNameCite{DeepJudge}{chenCopyRightTesting2022}
\createNameCite{DeepFoolF}{wangFingerprintingDeepNeural2021}
\createNameCite{FCAE}{lukasDeepNeuralNetwork2020}
\createNameCite{SSF}{heSensitiveSampleFingerprintingDeep2019}
\createNameCite{MetaV}{panMetaVMetaVerifierApproach2022}
\createNameCite{TAFA}{panTAFATaskAgnosticFingerprinting2021}

\usetikzlibrary { decorations.pathmorphing, decorations.pathreplacing, decorations.shapes, } 
\newcommand{\uwave}[1]{%
    \tikz[baseline=(todotted.base)]{
        \node[inner sep=1pt,outer sep=0pt] (todotted) {#1};
        \draw[decorate,decoration={snake,amplitude=1}] (todotted.south west) -- (todotted.south east);
    }%
}%
\newcommand{\udash}[1]{%
    \tikz[baseline=(todotted.base)]{
        \node[inner sep=1pt,outer sep=0pt] (todotted) {#1};
        \draw[dashed] (todotted.south west) -- (todotted.south east);
    }%
}%
\newcommand{\uline}[1]{%
    \tikz[baseline=(todotted.base)]{
        \node[inner sep=1pt,outer sep=0pt] (todotted) {#1};
        \draw (todotted.south west) -- (todotted.south east);
    }%
}%

\title{Queries, Representation \& Detection: The Next 100 Model Fingerprinting Schemes}
\author{
    Augustin Godinot\textsuperscript{\rm 1, 2, 3, 5},
    Erwan Le Merrer\textsuperscript{\rm 2},
    Camilla Penzo\textsuperscript{\rm 5},
    François Taïani\textsuperscript{\rm 1, 2, 3},
    Gilles Trédan\textsuperscript{\rm 4}
}
\affiliations{
    \textsuperscript{\rm 1}Université de Rennes, France,
    \textsuperscript{\rm 2}Inria, Rennes, France,
    \textsuperscript{\rm 3}IRISA/CNRS, Rennes, France, \\
    \textsuperscript{\rm 4}LAAS/CNRS, Toulouse, France,
    \textsuperscript{\rm 5}PEReN, Paris France\\
    augustin.godinot@inria.fr
}

\usepackage{booktabs}                   
\usepackage{multirow}                   
\usepackage{cleveref}                   
\nocite{*}                              
\newtheorem{proposition}{Proposition}   
\usepackage{siunitx}                    

\begin{document}
\maketitle

\begin{abstract}
	The deployment of machine learning models in operational contexts represents a significant investment for any organisation. Consequently, the risk of these models being misappropriated by competitors needs to be addressed. In recent years, numerous proposals have been put forth to detect instances of model stealing. However, these proposals operate under implicit and disparate data and model access assumptions; as a consequence, it remains unclear how they can be effectively compared to one another. Our evaluation shows that a simple baseline that we introduce performs on par with existing state-of-the-art fingerprints, which, on the other hand, are much more complex. To uncover the reasons behind this intriguing result, this paper introduces a systematic approach to both the creation of model fingerprinting schemes and their evaluation benchmarks. By dividing model fingerprinting into three core components -- Query, Representation and Detection (\QRD) -- we are able to identify $\sim100$ previously unexplored \QRD combinations and gain insights into their performance. Finally, we introduce a set of metrics to compare and guide the creation of more representative model stealing detection benchmarks. Our approach reveals the need for more challenging benchmarks and a sound comparison with baselines. To foster the creation of new fingerprinting schemes and benchmarks, we open-source our fingerprinting toolbox.
\end{abstract}

Companies devote considerable resources (i.e. manpower, funds and energy) to developing efficient and accurate machine learning (ML) models. 
Many of these models are then deployed in production on online platforms to solve a wide array of business-critical tasks (e.g. recommendations or predictions of all kinds). 
However, it is well understood that extraction attacks, or simply infrastructure leaks, can allow competitors to access the model architecture \cite{ohReverseEngineeringBlackBoxNeural2018}, weights \cite{carliniStealingPartProduction2024}, and hyperparameters \cite{wangStealingHyperparametersMachine2018}. 
From financial risks, when the attacker can provide the same functionality at a fraction of the cost, to integrity risks, when the attacker could use the stolen model as a step to craft adversarial examples, \emph{Model stealing attacks} pose great risks for the model developer. 

\begin{figure}
	\centering
	\includegraphics{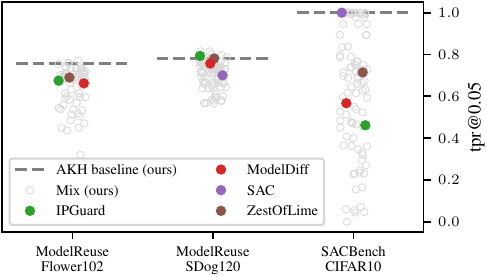}
	\caption{%
		The \tpr of most of the fingerprinting schemes proposed in the literature is
		at best as good as the simple baseline we introduce. Each colored dot represents the performance of an existing fingerprinting scheme evaluated on a given benchmark. The gray dots are fingerprinting schemes we created using our Query, Representation and Detection (\QRD) decomposition.}
    \label{fig:fingerprint_tpr}
\end{figure}

Although efforts have been devoted to defend models against extraction attacks \cite{294591,orekondyPredictionPoisoningDefenses2019,leeDefendingNeuralNetwork2019}, extraction defences have not yet been proven secure. 
Therefore, in addition to \emph{preventing} model stealing, companies need tools to \emph{detect} it. One such tool is \emph{model fingerprinting}. 
Similarly to how fingerprints can analyse the provenance of an image by identifying artefacts due to the compression scheme, the specific sensor technology, or even the up-scaling method \cite{ojhaUniversalFakeImage2023}, model fingerprints analyse the outputs of a ML model $h$ to extract artefacts that are characteristic of $h$ itself. 
In order to build a fingerprint for a given model $h$, the model owner first carefully selects a set of inputs $S$. 
The model owner then extracts a unique representation $Z_h$ from the output of their model $h$ when given $S$ as input. 
The representation $Z_h$ will serve as a fingerprint. 
The fingerprint $Z_h$ can later be compared with the fingerprint $Z_{h^\prime}$ extracted from the model $h^\prime$, which is suspected to be stolen. 
The fingerprint scheme depends on the input modality (e.g. text, image, or tabular data), on the model's task (e.g. classification, score, or recommendations), and hence on the domain of the output of $h$. 
In this work, as in most of the model fingerprinting literature, we consider image classification models. 
Note that, contrary to model watermarking methods, fingerprinting does not provide any theoretical guarantees on the false alarm rate (e.g. false positives). 
Thus, a strong empirical evaluation of model fingerprinting schemes is paramount to ensure their soundness in practice.

\paragraph{Problem} This paper presents a surprising artefact of fingerprinting evaluation. Fingerprinting evaluation consists in generating \emph{positive} and \emph{negative} model pairs $(h, h')$, where positive model pairs consist in a victim model $h$ and a model $h'$ stolen from $h$ (e.g. through model extraction), while for negative model pairs, $h$ and $h'$ are totally unrelated (e.g. trained on a different dataset). A collection of such positive and negative pairs is called \textit{benchmark}. \Cref{fig:fingerprint_tpr} displays the True Positive Rate (\tpr, see Paragraph \emph{Fingerprint evaluation} for the exact definition) of existing fingerprints on two existing benchmarks, \ModelReuse~and \SACBench. \Cref{fig:fingerprint_tpr} demonstrates that \emph{the simple baseline that we introduce (gray dashed lines) performs on par with existing state-of-the-art fingerprinting schemes (coloured dots), which are much more complex.}

In the following, we seek to understand the reasons behind this result by exploring the two key aspects of \Cref{fig:fingerprint_tpr}: How do existing fingerprints and benchmarks compare. Our contributions will be the following.
\begin{enumerate}
	\item We introduce a simple yet powerful baseline and provide theoretical guarantees on its performance. Albeit on a simple model copy detection task, this constitutes the first theoretical analysis of the guarantees of a model fingerprinting scheme.
	\item We survey and compare existing fingerprinting schemes for classification tasks. Our novel queries-representation-calibration decomposition (hereafter we coin \QRD) enables us to systematise and thus uncover new and unexplored fingerprinting schemes. 
    The novelty of \QRD lies in its mix of geometrical (distance between fingerprints leads to distance between models) and statistical insights (the fingerprint is then used to perform a statistical property test).
    \item We compare existing benchmarks and investigate their differences in both the way the pair of test models ($h, h'$) are generated and the distinguishability of the victim $h$ and suspected $h'$ models. Our work constitutes the first systematic comparison of classifier fingerprinting  benchmarks, and reveals insights into how to build more informative and challenging benchmarks. All the code required to re-run our experiments, implement new benchmarks and evaluate new fingerprints is available online\footnote{https://github.com/grodino/QuRD}.
\end{enumerate}

\section{Background and Setting}\label{sec:problem_setting}

\paragraph{Stealing ML models}
The possibilities for an adversary to steal a given model are endless. They could break into the infrastructure of their victim \cite{ben-sassonIsolationHallucinationHacking2024}, perform black-box model extraction attacks \cite{jagielskiHighAccuracyHigh2020,truongDataFreeModelExtraction2021} or just use the output of the victim's model to train their own. In this work, we consider adversaries seeking to steal the functionality of the victim's model.

\paragraph{Detecting IP violation via model fingerprinting}
The dominant approach to model fingerprinting is based on comparing the outputs of models on adversarial queries, as in \AFA, \TAFA, \IPGuard, \ModelDiff, \FUAP, \FCAE, \DeepFoolF, and \DeepJudge. Other approaches leverage the sensitivity of ML models at random points sampled from the train set (e.g. \SSF, \ModelGif), some explanations generated from the victim model $h$ \ZestOfLIME~or even train classifiers to distinguish stolen from benign model \MetaV. Some other works explore the use of natural images (images in the training/validation set) to craft their query set $S$, as in \FBI~or \SAC. All of these works try to detect model stealing, however comparison among them and the assumptions they make are rarely taken into consideration. In this work, we introduce a framework to compare and evaluate these fingerprints.

\paragraph{Problem setting} 
Consider an input space $\inputspace$, a space of labels $\outputspace = \set{1,\dots, C}$ with $C$ classes, a data distribution $\datad$ on $\inputspace$ and a ground truth concept $c \in \set{1, \dots, C}^\inputspace$. A first party called the \emph{victim} trains a model $h$ on a classification task $\mathcal{C}$, then deploys this model in production. A second party called the \emph{adversary} wishes to recreate a model $h'$ that is close to identical to $h$ ($h' \approx h$) to deploy it at a low cost.

The task of checking whether a \emph{suspected model} $h'$ is a copy of the \emph{victim model} $h$ is modeled as a property test \cite{goldreichIntroductionPropertyTesting2017}. A tester $\test$ is a (randomized) algorithm that takes two models $h$ and $h'$ as input and returns $1$ with high probability if $h'$ is stolen from $h$, $0$ else. 
\begin{equation*}
    \begin{array}{l r}
        \text{if } h = h',\; \probap{\test(h, h') = 1} > \frac{2}{3} & \textit{Copied model !} \\
        \text{if } h \neq h',\; \probap{\test(h, h') = 0} > \frac{2}{3} & \textit{Just an other model} \\
    \end{array}
\end{equation*}

The fingerprint (a.k.a. the property test) should be \emph{effective}, \emph{robust} and \emph{unique}. We also require the fingerprint to be \emph{efficient} in terms of queries and samples.
\begin{enumerate}
    \item \emph{Effectiveness}: if $h' = h$, then the suspected model is flagged by the victim with high probability.
    \item \emph{Robustness}: if $h'$ is a slightly modified version of $h$ (via fine-tuning, pruning, model extraction ...), then the suspected model should still be flagged.
    \item \emph{Uniqueness}: Original models $h' \neq h$ are not flagged.
    \item \emph{Efficiency}: the test uses few queries to the suspected model $h'$ and few samples $x$ from the data distribution.
\end{enumerate}

\paragraph{Accessibility of data and models}
The type of fingerprinting scheme that can be used by the victim depends on the access the victim has to the suspected model $h'$. We will assume that the victim can freely query the suspected model $h'$. Yet, the output of the suspected model will range from label-only query access, to top-K labels query access, probits or logits query access and even to gradients query access. Following the fingerprinting literature, it is assumed that the victim has full access to its training data and model $h$.

\section{Filling the gaps with the \baseline baseline}\label{sec:baseline}

The first contribution of this paper is the proposal and analysis of a simple yet powerful baseline, which, as we observed in \Cref{fig:fingerprint_tpr} performs at least as well as State-Of-the-Art fingerprinting schemes. 

It is assumed that the victim has access to samples from the input distribution, for example the test set they used to validate their model. 
The baseline refers to Tolstoy's Anna Karenina principle that states "All happy families are alike; each unhappy family is unhappy in its own way". 
Thus, instead of using random samples for the input space $\inputspace$, we look for points that are mis-classified by $h$ and compare the victim and suspected models on those points. 
Our baseline, coined the \textit{Anna Karenina Heuristic} (\baseline), proceeds as follows. 
First, the victim chooses a negative input: a point $x \sim \datad$ such that $h$ wrongly classifies $x$: $h(x) \neq c(x)$. 
We write $\snot{\datad_h}$ the resulting negative inputs distribution. 
Then, the victim queries the suspected model $h'$ on $x$. Finally, if $h'(x) = h(x)$ the suspected model $h'$ is flagged as stolen, otherwise $h'$ is deemed benign.

\begin{proposition}\label{prop:baseline_test}
    Consider $h, h' \in \outputspace^\inputspace$ two models and $\alpha = \probap{h(x) = c(x)}$ (resp. $\alpha' = \probap{h'(x) = c(x)}$) their accuracy. Let $\delta = d_H(h, h')$ be the relative Hamming distance between $h$ and $h'$ and $\delta_C = \probap{h(x) \neq h'(x) \given h(x) \neq c(x)}$. The property test $\test_b$ defined by \baseline enjoys the following guarantees:
    \begin{align}
        \text{If } h = h', \;
        &\probap[\datad]{\test_b(h, h') = 1} = 1 \\
        \text{If } h \neq h', \;
        &\probap[\datad]{\test_b(h, h') = 0} = \delta_C \geq \frac{\delta - (1 - \alpha')}{1 - \alpha}
    \end{align}
\end{proposition}

The proof of Proposition~\ref{prop:baseline_test} and the detailed algorithm can be found in the technical appendix. \Cref{prop:baseline_test} establishes that \baseline is a one-sided error test. Thus, in the favorable scenario where $h'$ is copied (i.e. not tampered with), $\test_b$ will always detect it. 
To simplify the analysis, we defined \baseline using only one query to the suspected model. To further decrease the False Negative Rate, one should run the baseline multiple times. A majority vote among the values returned by $\test_b$ decreases the False Negative Rate exponentially \cite{goldreichIntroductionPropertyTesting2017}.
If instead of selecting negative examples (points $x \in \inputspace$ that are wrongly classified by $h$), the victim was to use random samples according to $\datad$, the test would still have a one-sided error but the True Negative Rate $\probap[\datad]{\test(h, h') = 0}$ would be equal to the hamming distance $\delta$ between $h$ and $h'$. 
This gives us an idea on when \baseline can outperform schemes based on random sampling: either when the error rate $1 - \alpha$ of the victim model $h$ is low or when the error rate $1 - \alpha'$ of the suspected classifier $h'$ is low compared to $1 - \alpha$.

In practice, the \tpr of \baseline is displayed in \Cref{fig:fingerprint_tpr} in gray dashed lines.
On \ModelReuse~(SDog120 dataset) and on \SACBench, \baseline performs on par with the best existing fingerprints. 
On \ModelReuse~(Flower102 dataset), \baseline even performs better than the best existing fingerprints. 
In the two following sections we explore the reasons behind this observation by looking at the two players of \Cref{fig:fingerprint_tpr}: the fingerprints and the benchmarks used to compare them.

\section{Query, Representation \& Detection:\break the \QRD framework}\label{sec:qrd}

The literature on model fingerprinting does not provide a unified definition of model stealing detection. Most works focus on particular transformations of the stolen model, which they seek to detect. Only a few works \cite{caoIPGuardProtectingIntellectual2021, mahoFingerprintingClassifiersBenign2023, pengFingerprintingDeepNeural2022} are based on a mathematical formulation of the problem.
Some fingerprinting schemes (e.g. \ZestOfLIME~or \ModelGif) are described from a geometrical point of view: the goal is to create a distance between models to distinguish stolen models from unrelated models. On the other hand, some works are described from a statistical point of view: the goal is to test whether $h' = h$ or not. Thus, comparing and categorizing existing fingerprints is not trivial.
As a second contribution to this paper, we propose an original decomposition of the existing (and future) fingerprinting schemes into three core components: 
\begin{enumerate}
    \item \textbf{Query Sampling}, which generates the query set $S \subset \inputspace$ on which to query $h$ and $h'$, e.g. selecting a subset of the victim model training set $h$.
    \item \textbf{Representation}, which computes a compact representation $Z_h{=}g(Y_h)$ and $Z_{h'}{=}g(Y_{h'})$ of the answers $Y_h {=} \set{ h(x) {\sut} x \in S }$ and $Y_{h'} {=} \set{h'(x) {\sut} x \in S}$ that are returned by the two models $h$ and $h'$ on the sample $S$. A basic strategy is to use the raw answers as a representation, that is, $Z_h {=} Y_h, Z_{h'} {=} Y_{h'}$.
    \item \textbf{Detection}, which uses the two fingerprints $Z_h$ and $Z_{h'}$, and possibly a set of calibration fingerprints $\set{Z_i}_i$, to decide whether $h'$ is a stolen version of $h$ or not. 
\end{enumerate}

\begin{table*}[t]
    \centering
    \renewcommand{\arraystretch}{1.3}
    \begin{tabular}{c l  l p{5.2cm} l p{2cm} p{2.3cm}@{}}
        \toprule
        & & \textbf{Uniform} & \textbf{Adversarial} & \textbf{Negative} & \textbf{Subsampling} & \textbf{Joint detector training} \\
        \midrule
        \multirow{3}{*}{\rotatebox[origin=t]{90}{Seed set $S_\text{seed}$}} & input space & $\emptyset$ & \IPGuard & $\emptyset$ & $\emptyset$ & \uline{\MetaV} \\ 
        & test set & $\emptyset$ & \udash{\DeepJudge},  \FCAE & \FBI & $\emptyset$ & $\emptyset$ \\
        & train set & \uwave{\ModelGif} & \uline{\emph{\ModelDiff}}, \uline{\FUAP}, \IPGuard, \AFA, \uwave{\ModelGif}, \udash{\DeepJudge}, \udash{\SSF\footnotemark} & \udash{\textbf{\SAC}} & \uline{\textbf{\ZestOfLIME}}, \udash{\textbf{\SAC}}  & \uline{\FUAP} \\
        \bottomrule
    \end{tabular}
    \caption{Type of seed set $S_\text{seed}$ (rows), Query Sampling (Q) (columns), model access (emphasis) and Representation (R) (decorations) used. Adversarial sampling dominates the fingerprinting literature. Fingerprinting scheme appearing in multiple cells either require or can accomodate both Sampling/seed types. The text decoration stands for the access required to the remote suspected model $h'$: no decoration = label access, \uline{underline} = probits access, \udash{dashed underline} = label or probit access, \uwave{wavy underline} = gradients access. The text emphasis indicate the type of Representation: no emphasis = raw model outputs, \emph{italicized} = pairwise representation, \textbf{bold} = listwise representation. \footnotemark[1]\footnotesize \SSF~actually uses \emph{sensitive samples} instead of adversarial samples.}
    \label{tab:sampling_representation}
\end{table*}

\subsection{Query Sampling (Q)}
Existing approaches use four main techniques to build the query set when generating fingerprints, \emph{Uniform sampling}, \emph{Adversarial sampling}, \emph{Negative sampling}, and \emph{Subsampling} (see \Cref{tab:sampling_representation}). Query Sampling (Q) methods are based on the transformation of a seed query set $S_\text{seed}$, which is either the training set or the test set used by the victim when generating $h$ (both assumed to follow the same data distribution $\datad$), or images composed of random pixel values.

\subsubsection{Uniform sampling}
The easiest way to generate $S$ is to sample uniformly from the data distribution or from a seed set $S_\text{seed} \subset \inputspace$.%
\begin{equation}
    S \sim \datad \text{ or } S \sim \mathcal{U}(S_\text{seed}) 
\end{equation}

\subsubsection{Adversarial sampling} 
Adversarial sampling exploits the intuition that models tend to be characterized by their decision-boundary~\cite{lemerrerAdversarialFrontierStitching2020,caoIPGuardProtectingIntellectual2021,liModelDiffTestingbasedDNN2021}. Compared to uniform sampling, adversarial sampling leads to a better detection rate for a lower query budget $s$. Starting from a set of seed inputs $S_\text{seed} {\subset} \inputspace$, adversarial sampling computes a set of samples $S_\text{adv}$, targeted or not, using the following optimization procedure.%
\begin{equation}\label{eq:adv:sample}
    S_\text{adv} = \Big\{
        \argmax_{u, \norm{x - u} < \epsilon} d\big(h(x), h(u)\big), x \in S_\text{seed}
    \Big\}
\end{equation}
Common methods used for solving \Cref{eq:adv:sample} include Projected Gradient Descent~\cite{madryDeepLearningModels2018} or DeepFool~\cite{moosavi-dezfooliDeepFoolSimpleAccurate2016}. Finally, the final query set is the concatenation of the seed and adversarial samples $S = (S_\text{seed}, S_\text{adv})$. 

\subsubsection{Negative sampling}
As for adversarial sampling, negative sampling \cite{guanAreYouStealing2022} enjoys better detection rates for a given query budget. However, it does not need to compute gradients of $h$, it just needs query access to $h$, which can dramatically speed up the generation of the query set $S$. The core intuition follows that if $h'$ makes the same mistakes as $h$, there is a high probability that the adversary stole $h$. 
\begin{equation}
    S \subset S_\text{seed} \text{ subject to } \forall x \in S, h(x) \neq c(x) 
\end{equation}

\subsubsection{Subsampling} 
Subsampling exploits domain knowledge to create new samples $V(x){=} \set{x_j}_j$ in the vicinity of a seed point $x$. Compared to negative and adversarial sampling, subsampling allows to create a large query-set with few samples from the data distribution.
\begin{equation}
    S = (S_\text{seed},\set{V(x)}_{x \in S_\text{seed}}).
\end{equation}
\citeauthor{jiaZestLIMEArchitectureIndependent2022} uses the super-pixel sampling technique of LIME \cite{ribeiroWhyShouldTrust2016} to generate images around each image in a seed set $S_\text{seed}$.

\subsection{Representation (R)}
Once the model $h$ and $h'$ have been queried on a sample of data points, the resulting outputs $Y_h$ and $Y_{h'}$ must be recorded using some representation. We have identified three strategies in the literature: \emph{Raw Labels/Logits}, \emph{Pairwise correlation}, and \emph{Listwise correlation}.

\subsubsection{Raw labels/logits}
The simplest representation of the set of answers collected from the two models would be the set of answers themselves (labels or logits). However, depending on the way $h'$ was constructed (or not) from $h$, different representations are more suitable.
\begin{equation}
    Z_h = Y_h \in (\R^C)^s \text{ (logits) or } \set{1, \dots, C}^s \text{ (labels)}
\end{equation}

\subsubsection{Pairwise correlation} 
When the audit set $S$ consists of pairs of samples $(x, u)$ that have a specific meaning (e.g. $u$ is an adversarial version of $x$ as in \ModelDiff), it is interesting to use these pairwise comparisons as the representation of the model.
\begin{equation}
    Z_h = \left(d(h(x), h(u))\right)_{(x, u) \in S} \in \R^\frac{s}{2}
\end{equation}

\subsubsection{Listwise correlation} 
Generalizing the idea of pairwise correlation, if the audit samples are not specifically paired but comparison is still meaningful, the victim can compute the similarity between all pairs of answers and use the resulting similarity matrix as representation. This is what is used by \SAC.
\begin{equation}
    Z_h = \left(d(h(x), h(u))\right)_{x \in S, u \in S} \in \R^{s \times s}
\end{equation}

\subsection{Detection (D)}
Finally, once the victim has generated the fingerprints of their model and that of the suspected model ($Z_h$ and $Z_{h'}$), the last step is to compare $Z_h$ and $Z_{h'}$ to decide whether to flag $h'$ or not.

There exists two approaches to Detection (D): directly compute a distance (e.g. hamming as in \AFA~or mutual information as in \FBI) between the generated fingerprints or learn a classifier that takes the two fingerprints and outputs a theft probability score as in \MetaV. In both cases, the victim needs access to its own pool of fingerprints from unrelated models $\mathcal{G} = \set{G_{1},\ldots,G_{\abs{\mathcal{G}}}}$, to calibrate the detection threshold.

\subsection{The next 100 fingerprints}
In this subsection, we highlight the benefits of our novel \QRD decomposition for creating new and improved fingerprinting schemes and compare the existing fingerprints on a previously underexplored axis: the query budget. 

\paragraph{Fingerprint evaluation}
The \emph{Effectiveness}, \emph{Robustness} and \emph{Uniqueness} of fingerprints are evaluated by computing the Receiver-Operator Curve (ROC). The final Detection (D) step consists in thresholding a distance or the output of a classifier based on the fingerprints $Z_h$ and $Z_{h'}$. The ROC shows the relationship between the True Positive Rate (TPR), which is the proportion of positive pairs $(h, h')$ that are flagged as positive by the fingerprint, and the False Positive Rate (FPR), which is the proportion of negative pairs $(h, h')$ that are flagged as positive by the fingerprint. The ROC captures the trade-off between the cost to the victim of missing a stolen model compared to the cost of wrongly flagging a model as stolen. Recognizing the high cost of False Positives for the victim, we will report the TPR such that the FPR is below a threshold of $5\%$: \tpr, averaged over $5$ runs with independent random seeds.

\begin{figure}[t]
    \centering
    \includegraphics[width=\linewidth]{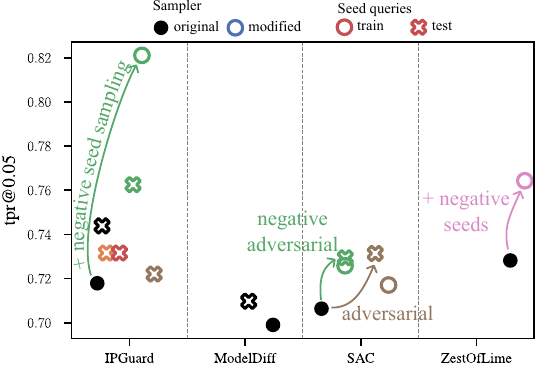}
    \caption{\tpr gains on \ModelReuse~obtained by modifying the sampler of existing fingerprints. The sampler can be modified in two ways: drawing seed queries from the train vs test set (materialized as circles vs crosses) or using a different queries sampler (materialized as a different color). Selecting negative seed inputs for adversarial generation instead of the original seeds can lead to improvements on the order of $10$ points ($+14\%$).}
    \label{fig:qrd_improvements}
\end{figure}

\paragraph{Creating new fingerprints using the \QRD framework}
Following our \QRD framework, \Cref{tab:sampling_representation} categorizes exiting fingerprints (listed previously in Background and Setting). 
\Cref{tab:sampling_representation} shows that a large part of the literature focused on fingerprints based on adversarial sampling. 
Several \QRD combinations have not been explored yet by the literature. 
Moreover, the schemes always focus on using only one type of Query Sampling (Q) but very rarely explore chaining or mixing, e.g.using negative samples as the seeds for generating adversarial examples. 
Thus, to explore the space of \QRD combinations, we reimplemented the Query Sampler, Representation, and Detection of four existing fingerprints: \ModelDiff, \SAC, \IPGuard~and \ZestOfLIME. 
We mixed them to create ${\sim}100$ new fingerprints. 
In \Cref{fig:fingerprint_tpr}, gray-edged dots represent such \QRD combinations. 
Of course, not all new combinations are worth considering, as many \QRD combinations exhibit lower \tpr than existing fingerprints. 
Thus, in \Cref{fig:qrd_improvements} we show the potential improvements that can be reached by modifying the Query Sampler (Q) and/or the seed set $S_\text{seed}$ of existing schemes on \ModelReuse. 
\Cref{fig:qrd_improvements} shows that it is possible to increase the \tpr of \IPGuard~by $10$ points ($+14\%$) simply by choosing negative seed samples as the starting points for the generation of adversarial examples.

\paragraph{Comparing apples to apples: a focus on the query budget}
\begin{table}[t]
    \centering
    \footnotesize
    \caption{Stealing and obfuscation methods implemented by different benchmarks.}
    \label{tab:bench_tasks}
    \begin{tabular}{@{}p{1cm} p{1cm} p{1.3cm} p{1.3cm} p{1.3cm}@{}}
        \toprule
         &  & ModelReuse Flower102 & ModelReuse SDog120 & SACBench CIFAR10 \\
        \midrule
        \multirow[c]{5}{1cm}{model leak} & same & \checkmark & \checkmark & \checkmark \\
         & quantize & \checkmark & \checkmark & $\times$ \\
         & finetune & $\times$ & $\times$ & \checkmark \\
         & transfer & $\times$ & $\times$ & \checkmark \\
         & prune & \checkmark & \checkmark & \checkmark \\
         \cmidrule(lr){1-5}
        \multirow[c]{4}{1cm}{Model extraction} & probits & \checkmark & \checkmark & \checkmark \\
        & label & \checkmark & \checkmark & \checkmark \\
        & adversarial (labels) & $\times$ & $\times$ & \checkmark \\
        \bottomrule
    \end{tabular}
\end{table}

\begin{figure}[t]
    \centering
    \includegraphics[width=\linewidth]{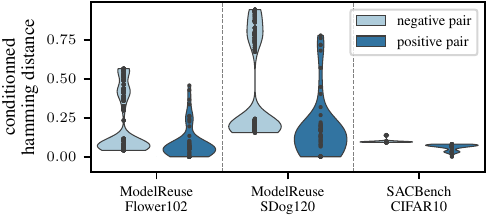}
    \caption{Distribution of the conditioned Hamming distance $d_C(h, h')$ between the models of each positive/negative $(h, h')$ pair.}
    \label{fig:conditioned_hamming}
\end{figure}
\begin{figure*}[t]
	\includegraphics{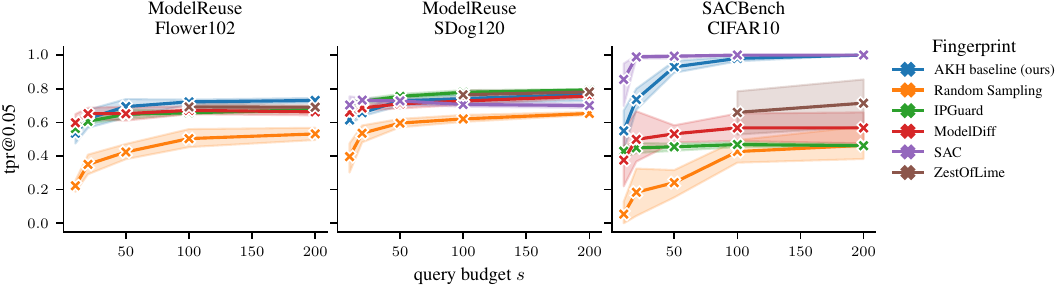}
	\caption{The effect of the query budget $s$ on the \emph{Efficiency} and \emph{Robustness} of existing fingerprints, as measured by \tpr.}
    \label{fig:tpr_vs_budget}
\end{figure*}

Although not displayed in \Cref{tab:sampling_representation}, the query budget required by the existing fingerprints can vary greatly. For example, \ZestOfLIME~requires from $\num{1000}$ to $\num{128000}$ queries while \FBI~only requires $\sim100$ queries to reach the advertised performance. In \Cref{fig:tpr_vs_budget} we show the \tpr of existing fingerprints along our \baseline baseline and selected \QRD variations. Keeping a small query budget is of paramount importance, mainly to remain stealthy against potential defenses \cite{oliynykKnowWhatYou2023}, but also to avoid disrupting the remote service with (tens to hundreds of) thousands of queries. Once more, we observe that fingerprints based on negative sampling equal or outperform fingerprints based on adversarial sampling. From $0$ to $100$ queries for \SACBench~and $0$ to $50$ for \ModelReuse, most fingerprints exhibit notable improvements at each query budget increment. After $100$ (or $50$) queries, most fingerprints show a plateau. Thus, it appears that there exists an optimal query budget, dependent on the benchmark but not on the fingerprinting scheme. Finally, schemes based on negative sampling appear to suffer a lower variance than adversarial-based fingerprints, especially on \SACBench.

Although the performance of most fingerprints plateau after $50$-$100$ queries, the performance of some fingerprints (e.g. \ModelDiff~and \SAC) suffers when the query budget increases from $100$ to $400$ queries. This phenomenon is observable only for schemes whose representations are based on a pairwise or a listwise comparison. We believe that when the number of query points is increased, the self-correlation increases regardless of the fact that a pair is positive or negative. Thus, the gap between the positive pair distance and the negative pair distance decreases with budget, which in turn decreases the performance of the fingerprint. 

\section{Fingerprinting benchmarks}\label{sec:benchmarks}

Because there are no strong guarantees regarding \emph{Effectiveness} and \emph{Robustness} of fingerprinting schemes, proper empirical evaluation is critical to assessing their performance. The main difficulty of evaluation lies in the definition (and implementation) of realistic \emph{positive} ($h'{=}h$) and \emph{negative} ($h'{\neq} h$) model pairs. To do this, we need to separate how the adversary steals the model (how to achieve $h'{=}h$) and how the adversary tries to conceal their theft by modifying the stolen model to avoid detection by the victim).

\paragraph{Stealing a model} 
\textbf{1) Model leak}: the adversary directly steals the architecture and weights of the model $h$ and uses them to solve the same task. This can happen via an internal leak \cite{franzenMistralCEOConfirms2024} or an attack on the company infrastructure \cite{ben-sassonIsolationHallucinationHacking2024}. 
\textbf{2) (Adversarial) model extraction} The adversary only has query access to the source model and trains their model based on the probits or the labels of the source model. 
The model extraction can either be probits or labels-based \cite{jagielskiHighAccuracyHigh2020,truongDataFreeModelExtraction2021}. 
In addition, depending on the threat model, the architecture trained by the attacker is not always the same as the victim model $h$ and the adversary might not have access to samples from the input domain \cite{truongDataFreeModelExtraction2021}.

\paragraph{Stolen model obfuscation} 
Once an attacker has stolen the model $h$, they will try obfuscating their model to hide their theft. 
To avoid detection by model fingerprinting, the adversary may act on a combination of three aspects of the model inference process. 
\textbf{1)~Model/weights tampering} As first approach, the adversary can directly modify the model itself to remove potential watermarks embedded in the weights of the model: weights pruning \cite{liuFinePruningDefendingBackdooring2018,liPruningFiltersEfficient2017}, model quantization and finetuning or transferring the model to a small private dataset \cite{liModelDiffTestingbasedDNN2021}.
\textbf{2)~Input modifications} The second concealment trick is to apply transformations to the inputs fed to the model to limit the effect of adversarial inputs \cite{mahoFingerprintingClassifiersBenign2023}: JPEG compression, equalization, or posterization. 
\textbf{3)~Output noise}: Finally, to avoid giving away too much information, the adversary can try to slightly alter the outputs of the model, e.g. returning only the Top-K labels, averaging the outputs over a neighbourhood of the input \cite{cohenCertifiedAdversarialRobustness2019} or implementing model-stealing defences \cite{294591,orekondyPredictionPoisoningDefenses2019}.

\subsection{The majority of benchmarked tasks are solved}\label{subsec:tasks_perfs}

\begin{table*}[t]
    \footnotesize
    \centering
    \caption{$\text{TPR}@0.05$ of the existing fingerprints with a budget of $100$ queries. For each task, the best performance are highlighted.}
    \label{tab:task_perf}
    \begin{tabular}{l|lllll | l|ll}
        \toprule
         & \multicolumn{5}{l}{Model leak} & \multicolumn{1}{l}{Probit extraction} & \multicolumn{2}{l}{Label extraction} \\
        Fingerprint & same & quantize & finetune & transfer & prune & vanilla & vanilla & adversarial \\
        \midrule
        IPGuard & \bfseries 1.0 ± \scriptsize0 & \bfseries 1.0 ± \scriptsize0 & \bfseries 1.0 ± \scriptsize0 & \bfseries 1.0 ± \scriptsize0 & \bfseries 0.94 ± \scriptsize.01 & 0.64 ± \scriptsize.02 & 0.12 ± \scriptsize0 & 0.02 ± \scriptsize.01 \\
        ModelDiff & \bfseries 1.0 ± \scriptsize0 & \bfseries 1.0 ± \scriptsize0 & \bfseries 1.0 ± \scriptsize0 & \bfseries 1.0 ± \scriptsize0 & \bfseries 0.94 ± \scriptsize.01 & 0.59 ± \scriptsize.05 & 0.14 ± \scriptsize.02 & 0.16 ± \scriptsize.07 \\
        Random Sampling & \bfseries 1.0 ± \scriptsize0 & 0.93 ± \scriptsize.03 & \bfseries 1.0 ± \scriptsize0 & 0.48 ± \scriptsize.2 & 0.71 ± \scriptsize.02 & 0.46 ± \scriptsize.01 & 0.07 ± \scriptsize.02 & 0.06 ± \scriptsize.05 \\
        SAC & \bfseries 1.0 ± \scriptsize0 & \bfseries 1.0 ± \scriptsize0 & \bfseries 1.0 ± \scriptsize0 & \bfseries 1.0 ± \scriptsize0 & 0.92 ± \scriptsize0 & \bfseries 0.81 ± \scriptsize0 & \bfseries 0.59 ± \scriptsize.02 & \bfseries 1.0 ± \scriptsize0 \\
        ZestOfLime & \bfseries 1.0 ± \scriptsize0 & \bfseries 1.0 ± \scriptsize0 & \bfseries 1.0 ± \scriptsize0 & 0.78 ± \scriptsize.17 & 0.86 ± \scriptsize0 & 0.74 ± \scriptsize.02 & 0.38 ± \scriptsize.05 & 0.29 ± \scriptsize.11 \\
        \baseline (ours) & \bfseries 1.0 ± \scriptsize0 & \bfseries 1.0 ± \scriptsize0 & \bfseries 1.0 ± \scriptsize0 & \bfseries 1.0 ± \scriptsize0 & 0.91 ± \scriptsize.01 & 0.78 ± \scriptsize.01 & 0.46 ± \scriptsize.01 & 0.92 ± \scriptsize.03 \\
        \bottomrule
    \end{tabular}
\end{table*}

The performance shown previously in \Cref{fig:fingerprint_tpr,fig:conditioned_hamming,fig:tpr_vs_budget} were all aggregated at a benchmark level. In this section, we separate the performance of the fingerprints with respect to the model-stealing and obfuscation methods. We will seek to answer the question \textit{What type of stealing and obfuscation methods can be considered as resolved issues and, hence, on which ones should practitioners focus?} Positive pairs are grouped by task, i.e., how the copied model $h'$ was created from $h$, along with their corresponding negative pairs. Each task corresponds to the combination of a stealing and an obfuscation method. 
This decomposition is especially interesting since, as we will observe, a large portion of the tasks are solved by all the fingerprints, while the rest, and more complicated tasks, allows to discriminate the different fingerprints much more clearly.

As for benchmark-aggregated performance discussed in the \QRD Section, \Cref{tab:task_perf} shows that \baseline is on par or surpasses all the previously introduced schemes. More interestingly, \Cref{tab:task_perf} reveals that a large part of the tasks considered by \ModelReuse~and \SACBench~(namely the same, quantization, finetuning, and transfer tasks) are completely solved by existing fingerprints, as well as by \baseline. The remaining unsolved tasks consist of model stealing by model extraction, using no obfuscation attempts. Surprisingly, adversarial label extraction is easily detected by fingerprints based on negative sampling but not by adversarial, random, or subsampling-based fingerprints. 
Model extraction detection is, thus, a hard subtask of model stealing detection.

The results of \Cref{tab:task_perf} highlight an issue with the current benchmarks: trying to detect if a suspected model $h'$ is the same as the victim's $h$ up to small model perturbations (pruning, quantization, etc.) is fundamentally different from detecting model extraction. These two objectives differ in difficulty to be detected (as we mentioned earlier), but they also differ greatly in the efforts the adversary has to consent to reach the same accuracy.

\subsection{Why does SACBench look so easy?}

As we observed in \Cref{fig:fingerprint_tpr}, the performance of fingerprints varies greatly from one benchmark to another. In this section, we try to uncover the reasons for this variability.
A fingerprinting benchmark is essentially a procedure to generate positive and negative model pairs $(h, h')$ by varying the model stealing and obfuscation methods. In the following, we investigate the properties of positive and negative pairs for each benchmark, in order to better understand the reasons why the various benchmarks seem to be unable to discriminate proposed fingerprint schemes and are beaten by the simple baseline presented in the previous section. \ModelReuse~and \SACBench~employ the same set of model stealing and obfuscation methods with two exceptions: \ModelReuse~uses model quantization as an obfuscation strategy, while \SACBench~performs adversarial model extraction. This explains the inferior performance of fingerprints based on adversarial sampling (\ModelDiff~and \IPGuard) on \SACBench.

However, the slight choice difference of the stealing and obfuscation methods included in \ModelReuse~compared to \SACBench~does not explain the exceptional performance of \baseline and \SAC~compared to the other fingerprints. To that end, in \Cref{fig:conditioned_hamming} we show the value of the conditioned Hamming distance $\delta_C$ (see \Cref{prop:baseline_test}) for all model pairs $(h, h')$.
We note that the variability of the distance between $h$ and $h'$ is much higher for \ModelReuse~than for \SACBench. This indicates that \SACBench's process for creating the positive and negative pairs may not introduce enough diversity in the generated models, which could lead to overestimating the performance of its fingerprints. However, as observed in \Cref{fig:fingerprint_tpr}, except \SAC, all fingerprints have a comparable \tpr on \SACBench~and \ModelReuse. 
To explain the difference in performance of \baseline and \SAC, we need to consider the separation between the distribution of $\delta_C(h, h')$ for the positive and negative model pairs $(h, h')$. \Cref{fig:conditioned_hamming} shows a better separation between $\delta(h, h')$ for positive and negative pairs in \SACBench. On the other hand, both datasets of \ModelReuse~show a large overlap in the distributions of distances of positive and negative pairs. Thus, since \SAC~is based on negative sampling, it appears that the generated positive and negative pairs of \SACBench~are especially well suited to the \SAC~fingerprint they introduce.

\section{Related works}\label{sec:related_works}
\textbf{Model-theft proactive defenses} An alternative to fingerprinting is for the victim to choose a proactive solution consisting in \emph{watermarking} their model (see, e.g., \cite{boenischSystematicReviewModel2021, regazzoniProtectingArtificialIntelligence2021} for an overview), or by defending it using defenses implemented at training or inference time \cite{oliynykKnowWhatYou2023, 294591}.

\noindent\textbf{Connections with tampering detection} A problem closely related to model fingerprinting is \emph{tampering} detection. The goal is to detect if a model served by a platform is the intended model originally sent by the owner, or if the model has been tampered with \cite{lemerrerTamperNNEfficientTampering2019,heSensitiveSampleFingerprintingDeep2019}, by backdoor attacks \cite{guBadNetsIdentifyingVulnerabilities2019} for instance.

\noindent\textbf{Connections with interpretable model distance} To debug model creation and to help ML audits, a body of work is interested in \emph{interpretable} model distances. Instead giving a single distance value, it also gives an explanation such as domains on where the models differ the most \cite{ridaDynamicInterpretabilityModel2023} or a simple approximation of the difference of the two models \cite{nairWhatChangedInterpretable2021}.

\section{Conclusion}\label{sec:conclusion}

Our systematic analysis of the existing model fingerprinting schemes and benchmarks revealed a concerning evaluation artifact: the benchmarks studied are either not discriminative or solved by our simple \baseline baseline.
Firstly, most tasks are solved with almost any fingerprint. Secondly, the created victim/stolen model pairs are too easy to distinguish from victim/benign model pairs.
Moreover, our \QRD framework reveals that schemes based on adversarial sampling are brittle compared to schemes using natural images.

While some of the tasks of model stealing detection can now be considered solved, several open challenges remain. One key issue is ensuring the robustness of fingerprinting techniques against adaptive adversaries who may actively attempt to evade detection. Furthermore, the development of effective fingerprints for other modalities than images would require further exploration.

\section{Acknowledgments}
The authors acknowledge the support of the French Agence Nationale de la Recherche (ANR), under grant ANR-24-CE23-7787 (project PACMAM). 
This project was provided with computing AI and storage resources by GENCI at IDRIS thanks to the grant AD011015350 on the supercomputer Jean Zay's V100 partition.
This research work was partially supported by the Hi! PARIS Center.
A.G. would like to thank Dimitrios Los for the fruitful discussions on the theoretical analysis.

\begin{links}
    \link{Code}{https://github.com/grodino/QuRD}
\end{links}

\bibliography{Godinot}

\appendix

\section{Proof of Proposition~\ref{prop:baseline_test}}\label{apx:sec:baseline_proof}
\begin{algorithm}[ht]
    \caption{The proposed baseline: \baseline}
    \label{alg:baseline}
    \baseline($\datad$, whitebox $h$, query access $h'$)
    
    \begin{algorithmic}[1]
        \STATE Sample $x \sim \datad$ such that $h(x) \neq c(x)$
        \IF {$h(x) = h'(x)$} 
        \STATE \textbf{Return} $1$ (\texttt{Stolen})
        \ENDIF
        \STATE \textbf{Return} $0$ (\texttt{Benign})
    \end{algorithmic}
\end{algorithm}

\begin{proof}
    \underline{Case $h = h'$}
    In this case, $\forall x \in \inputspace, h(x) = h'(x)$. Thus, $\test$ will always return $1$.

    \underline{Case $h \neq h'$}
    \begin{align*}
        \probap{\test^\datad(h, h') = 0} =& \probap[x \sim \snot{\datad_h}]{h(x) \neq h'(x)} \\
        =& \probap{h(x) \neq h'(x) \given h(x) \neq c(x)} \\
        =& \frac{\probap{h(x) \neq h'(x), h(x) \neq c(x)}}{\underbrace{\probap{h(x) \neq c(x)}}_{1 - \alpha}} 
    \end{align*}

    We now decompose the event $h(x) \neq h'(x)$ on the partition $(h(x) = c(x), h(x) \neq c(x))$.
    \begin{align*}
        \probap{h(x) \neq h'(x)} =& \probap{h(x) \neq h'(x), h(x) \neq c(x)} \\
        &+ \probap{h(x) \neq h'(x), h(x) = c(x)} 
    \end{align*}

    Using the inclusion $\set{h(x) \neq h'(x), h(x) = c(x)} \subset \set{h'(x) \neq c(x)}$
    \begin{equation*}
        \probap{h(x) \neq h'(x), h(x) = c(x)} \leq \underbrace{\probap{h'(x) \neq c(x)}}_{1 - \alpha'}
    \end{equation*}

    Thus, 
    \begin{align*}
        &\probap{\test^\datad(h, h') = 0} \\
        &= \frac{\probap{h(x) \neq h'(x), h(x) \neq c(x)}}{\probap{h(x) \neq c(x)}} \\
        &= \frac{\overbrace{\probap{h(x) \neq h'(x)}}^\delta - \overbrace{\probap{h(x) \neq h'(x), h(x) = c(x)}}^{\leq 1 - \alpha'}}{\underbrace{\probap{h(x) \neq c(x)}}_{1 - \alpha}} \\
        &\geq \frac{\delta - (1 - \alpha')}{1 - \alpha}.
    \end{align*}
\end{proof}

\section{Evaluation Setup}
The fingerprints we re-implemented are \IPGuard, \ModelDiff, \SAC~and \ZestOfLIME. We based our implementation on the descriptions of the schemes in their respective papers and re-used part of the authors' code when available.
We choose two benchmarks -- \ModelReuse~and \SACBench-- spanning three common vision datasets: Stanford Dogs \cite{KhoslaYaoJayadevaprakashFeiFei_FGVC2011}, Oxford Flowers \cite{nilsbackAutomatedFlowerClassification2008} and CIFAR10 \cite{krizhevskyLearningMultipleLayers2009} which we abbreviate as SDog120, Flower102 and CIFAR10. We used the model weights released by the authors of the respective benchmarks. For each experiment, we report the average (and standard deviation) over five runs for each setting. The experiments were run on a compute cluster. The nodes were based on an Intel Cascade Lake 6248 processor with 16Go Nvidia Tesla V100 SXM2 GPUs. The code is available at the following anonymized repository: \url{https://anonymous.4open.science/r/aaai25-E33E/}.

\section{Details on the computation of the True and False Positive Rate}
The True Positive Rate and False Postive Rate are computed as follows. Consider a fingerprint (as defined in the problem setting section) $\test: \left(\outputspace^\inputspace, \outputspace^\inputspace\right) \to \set{0, 1}$. Define $\mathbb{V}$ to be a set of victim models and for each victim model $h \in \mathbb{V}$, $\mathbb{S}(h)$ is a set of models stolen from $h$ and $\mathbb{U}(h)$ is a set of models unrelated to $h$. A benchmark is a triplet $\mathbb{B} = \left(\mathbb{V}, (\mathbb{S}(h))_{h \in \mathbb{V}}, (\mathbb{U}(h))_{h \in \mathbb{V}} \right)$. The True and False positive Rate reported in the paper are computed as follows.
\begin{align}
    \text{TPR}(\mathbb{B}) = \frac{1}{\card{\mathbb{V}}} 
        \sum_{h \in \mathbb{V}} \frac{\sum_{h' \in \mathbb{S}(h)} \indb{\test(h, h') = 1}}{\card{\mathbb{S}(h)}} \\
    \text{FPR}(\mathbb{B}) = \frac{1}{\card{\mathbb{V}}} 
        \sum_{h \in \mathbb{V}} \frac{\sum_{h' \in \mathbb{U}(h)} \indb{\test(h, h') = 1}}{\card{\mathbb{U}(h)}}        
\end{align}

\end{document}